# Automated Segmentation of the Optic Disk and Cup using Dual-Stage Fully Convolutional Networks

*Lei Bi[1], Yuyu Guo[2], Qian Wang[2], Dagan Feng[1,3], Michael Fulham[1,4,5], Jinman Kim[1]*
[1]School of Information Technologies, University of Sydney, Australia
[2]School of Biomedical Engineering, Shanghai Jiao Tong University, China
[3]Med-X Research Institute, Shanghai Jiao Tong University, China
[4]Department of PET and Nuclear Medicine, Royal Prince Alfred Hospital, Australia
[5]Sydney Medical School, University of Sydney, Australia

## ABSTRACT

Automated segmentation of the optic cup and disk on retinal fundus images is fundamental for the automated detection / analysis of glaucoma. Traditional segmentation approaches depend heavily upon hand-crafted features and a priori knowledge of the user. As such, these methods are difficult to be adapt to the clinical environment. Recently, deep learning methods based on fully convolutional networks (FCNs) have been successful in resolving segmentation problems. However, the reliance on large annotated training data is problematic when dealing with medical images. If a sufficient amount of annotated training data to cover all possible variations is not available, FCNs do not provide accurate segmentation. In addition, FCNs have a large receptive field in the convolutional layers, and hence produce coarse outputs of boundaries. Hence, we propose a new fully automated method that we refer to as a dual-stage fully convolutional networks (DSFCN). Our approach leverages deep residual architectures and FCNs and learns and infers the location of the optic cup and disk in a step-wise manner with fine-grained details. During training, our approach learns from the training data and the estimated results derived from the previous iteration. The ability to learn from the previous iteration optimizes the learning of the optic cup and the disk boundaries. During testing (prediction), DSFCN uses test (input) images and the estimated probability map derived from previous iterations to gradually improve the segmentation accuracy. Our method achieved an average Dice co-efficient of 0.8488 and 0.9441 for optic cup and disk segmentation and an area under curve (AUC) of 0.9513 for glaucoma detection.

*Index Terms*— Segmentation, Fully Convolutional Networks (FCN), Glaucoma

## 1. INTRODUCTION

Glaucoma is the second leading cause of blindness [1]. It is usually caused by elevated intraocular pressure [2], which causes mechanical damage to the optic nerve, which contains the retinal nerve fibers. A retinal fundus scan is commonly used to detect or track the progress of glaucoma [3]. It is a laser beam scan of the fundus, allowing the evaluation of structural relationships of the retina, optic disk and cup [4]. The cup-to-disk ratio (CDR) compares the diameter of the optic cup to the disk and partially represents disease status [5, 6]. Previous studies [7, 8] show that a larger vertical CDR is associated with glaucoma progression. Although the CDR is not diagnostic it is useful in clinical practice for the evaluation of glaucoma [7, 8]. The manual determination of CDR is, however, subjective, operator dependent, poorly reproducible and time-consuming. For these reasons, there has been increasing research interest to develop an operator-independent approach.

Traditional segmentation methods, based on shape models, graph cut and superpixels rely on hand-crafted features and a priori knowledge of the user who refines the segmentation [9, 10, 11]. In addition, these methods usually depend on low-level features, such as local texture that do not capture image-wide variations. Furthermore, the performance of such methods relies on correctly tuning a large number of parameters that restrict broad application.

Deep learning methods based on fully convolutional networks (FCNs) have recently shown success in segmentation problems [12]. This is primarily attributed to FCNs being able to leverage large datasets to learn a feature representation that combines low-level appearance information in lower layers with high-level semantic information in deeper layers [12]. In addition, FCNs can be trained end-to-end manner for efficient inference, i.e., images are taken as inputs and the segmentation results are directly outputted. The dependence of FCNs on large amounts of annotated data are problematic with medical images because there is a scarcity of annotated training data and so the application of FCNs can results in coarse detection and poor definition of the boundaries.

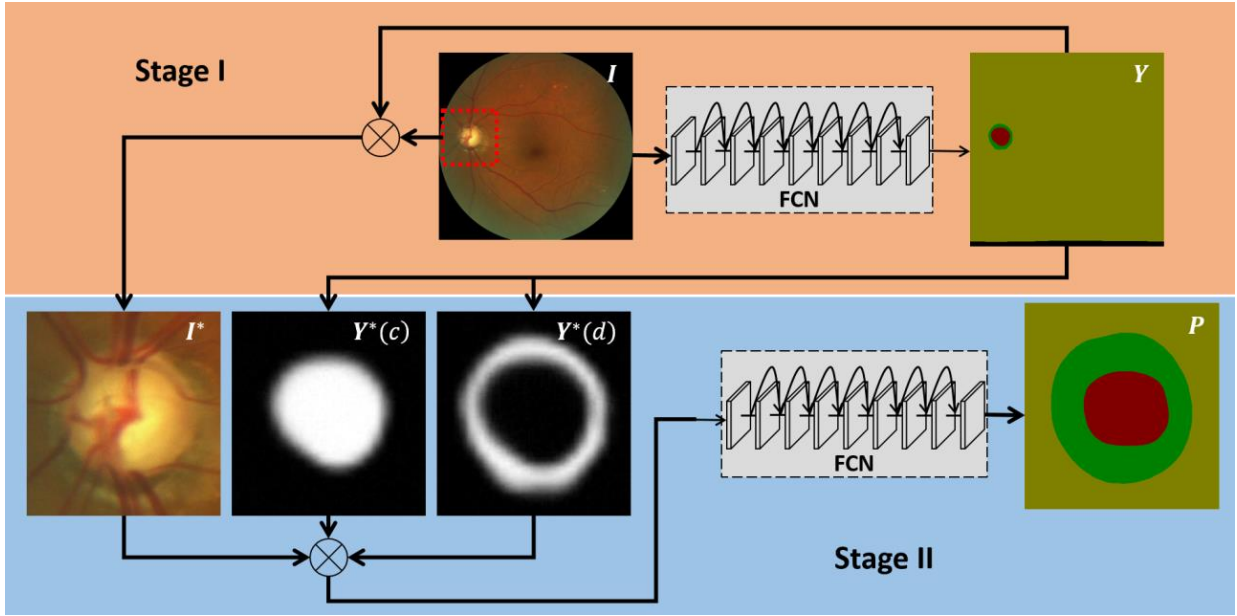

Figure 1. Flow diagram of our dual-stage fully convolutional networks (DSFCN).

Hence, we propose a dual-stage FCN that iteratively refines and constrain the boundaries at the training and testing phases. During training, our DSFCN learns from the data and the estimated derivations from the previous iteration, and optimizes the learning of the boundaries of the optic cup and disk. During testing (prediction), the DSFCN uses test images and the estimated probability map derived from the previous iterations to improve segmentation.

## 2. METHODS AND MATERIALS

### 2.1. Materials

We used the MICCAI 2018 – Retinal Fundus Glaucoma Challenge (REFUGE) dataset (http://refuge.grand-challenge.org) that contains annotated images of the retinal fundi, acquired on a variety of devices at international clinical centers. The training dataset has 400 images (40 glaucoma and 360 non-glaucoma) with a resolution of 2124×2056 pixels. For our evaluation, we randomly split the training images into 360 training and 40 validation images. An additional 400 images provided by the challenge organizer were used as the test set. The ground truth for the test set was not available to the public.

### 2.2. Pre-processing

We padded the image short-axis with zeros and then resized the image into 512×512 pixels to keep the original aspect ratio. In addition, as reflected in current literature [13], we used the green channel of the image, because the blood vessels that converge on the disk have the highest contrast in the green channel, for training and testing the segmentation network.

### 2.3. Fully Convolutional Networks (FCNs)

The FCN architecture was converted from convolutional neural networks (CNNs) for efficient dense inference [12]. The FCNs contain downsampling and upsampling components. Downsampling has stacked convolutional layers to extract high-level semantic information and has been routinely used in CNNs for image classification tasks [14]. Upsampling has stacked deconvolutionals, which are transposed convolutional layers that upsample the feature maps derived from downsampling element to output the segmentation results [15]. The FCN architecture can be trained end-to-end, in medical image segmentation, by minimizing the overall loss function (e.g., cross-entropy loss) between the predicted labels and the ground truth annotation of the training data. The network for segmentation can be defined as:

$$Y = U_s(F_s(I, \omega), \varphi) \qquad (1)$$

where $Y$ is the output prediction, $I$ is the input image, $F_S$ denotes the feature map produced by the stacked convolutional layers with a list of stride values $S$. $U$ represents the upsampling process to scale the feature up into the original input size. $\omega$ and $\varphi$ denotes the learned parameters. The FCN parameters (weights) can then be updated iteratively using e.g. a stochastic gradient descent (SGD) algorithm.

## 2.4. Dual-Stage Fully Convolutional Networks (DSFCN)

Our DSFCN embeds the probability map produced at the previous FCN for training and testing (shown in Fig. 1). We also cropped out the detected optic disk and cup area to reduce the reception field. Then the calculation can be defined as:

$$P = U_s(F_s((I^*; Y^*(c); Y^*(d)), \omega), \varphi) \qquad (2)$$

where $P$ is the final output of the prediction of the DSFCN, $Y(c)$ denotes the probability map derived from previous FCN for predicting the optic cup and $Y(d)$ for the optic disk. The cropped function is identified by *, where the cropped region only encapsulates the area within optic disk. We based our FCN architecture on a conventional 101-layer residual network (ResNet) [16, 17].

## 2.5. Glaucoma Classification

For glaucoma classification, we firstly calculate the ratio of optic cup to disk, which is defined as:

$$\gamma = \frac{|P(c)|}{|P(d)|} \qquad (3)$$

Where |·| represents the number of pixels within the segmented regions. Then, we used a curve fitting approach to find a sigmoid curve that can map the ratio $\gamma$ to the ground truth labels, where glaucoma is labelled as 1 and non-glaucoma as 0.

## 2.6. Implementation Details

Current literature suggests fine-tuning for medical images [20], where the lower layers of the fine-tuned network are more general filters (trained on general images) and the higher layers are more specific to the target problem. Therefore, we trained the DSFCN and fine-tuned both stages with a pre-trained model trained on the ImageNet dataset [18] for 200 epochs using a fixed learning rate of 0.0005. Data argumentation including random scaling, crops and flips that were used to further improve the robustness of the model. The training image batch size was set to 1 and the training process took approximately 35 hours to train a single stage on a single 12GB Titan X GPU (Maxwell architecture).

## 3. RESULTS AND DISCUSSION

### 3.1. Materials and Experimental Setup

The submission system only allowed two submissions, so we used the validation set to compare our method to current FCN methods based on VGGNet (16-layer) [12], U-NET [19] and ResNet (101-layer) [17] architectures and used the test set to evaluate our method at individual stages. We used the common segmentation evaluation metrics for comparison including: the dice similarity coefficient (DSC), the Jaccard Index (Jac.), sensitivity (Sen.), specificity (Spe.), and accuracy (Acc.). The submission system used additional segmentation metrics including mean absolute error for cup to disk ratio (MAE - CDR) and glaucoma detection metrics including area under curve (AUC), for comparison and ranking.

**Table 1:** Segmentation results of our method compared to other methods for the validation dataset, where **Bold** represents the best results.

|  |  | VGGNet | U-NET | ResNet | DSFCN |
|---|---|---|---|---|---|
| **Optic Cup** | Dice | 0.8513 | 0.8592 | 0.8652 | **0.8677** |
|  | Jac. | 0.7479 | 0.7602 | 0.7655 | **0.7708** |
|  | Sen. | 0.8671 | **0.9072** | 0.8923 | 0.8476 |
|  | Spe. | 0.9993 | 0.9993 | 0.9994 | **0.9996** |
|  | Acc. | 0.9987 | 0.9988 | 0.9982 | **0.9989** |
| **Optic Disk** | Dice | 0.8419 | 0.8720 | 0.8472 | **0.8733** |
|  | Jac. | 0.7363 | 0.7785 | 0.7445 | **0.7829** |
|  | Sen. | 0.8381 | **0.8766** | 0.8155 | 0.8705 |
|  | Spe. | 0.9983 | 0.9985 | **0.9988** | 0.9986 |
|  | Acc. | 0.9962 | 0.9969 | 0.9965 | **0.9970** |

**Table 2:** Segmentation and classification results for the online test dataset.

|  | Segmentation | | | Classification | | Overall Rank |
|---|---|---|---|---|---|---|
|  | Dice – Optic Cup | Dice – Optic Disk | MAE – CDR | AUC | Sen. |  |
| **Stage I** | 0.8107 | 0.9350 | 0.0816 | 0.9383 | 0.8250 | - |
| **Stage II** | 0.8488 | 0.9441 | 0.0471 | 0.9513 | 0.8992 | 4 / 28 |

### 3.3. Results

Table 1 and Fig. 2 show that our DSFCN performed better when compared with the current FCN methods based on VGGNet (16-layer) [12], U-NET [19] and ResNet (101-layer) [17] architectures. We suggest that the slight improvement relates to the dual-stage approach. Fig. 2 presents one example study where our method improved the segmentation of the optic cup (annotated in black), where all other methods including VGGNet, U-NET and ResNet suffered from over segmentation.

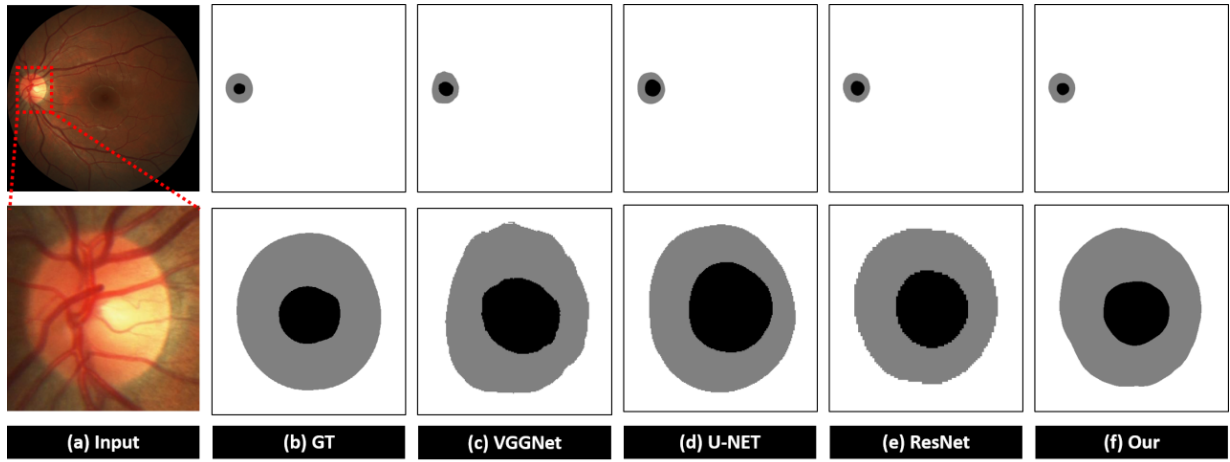

Figure 2. Segmentation results from one example. Top left is an image of the optic fundus with the yellow 'full-moon shaped' optic disk apprearing 'yellow'; vessels converging on the disk are red. (a) input images, where the bottom row is the cropped view of the top row; (b) ground truth (GT) annotation; (c-e) segmention results for VGGNet, U-NET, ResNet; and (f) for our method with optic cup in black, optic disk in grey and retina in white.

Table 2 shows the segmentation and classification results of our method at individual stages on the test dataset. The dual-stage learning improves the segmentation of the disk and the cup, which also helps the detection of glaucoma. Table 2 also shows that when our method is coupled to a conventional 101-layer ResNet architecture, the performance is competitive (4 out of 28 teams) without any additional manipulations. This finding suggests that our method may continue to improve with more convolutional layers, data-specific cost functions and post-processing techniques.

## 4. CONCLUSION

We outline an automated segmentation method for images of the optic fundi where we use a dual-stage approach to learn and infer the location of the optic disk and cup. Our experiments with the MICCAI 2018 REFUGE challenge dataset show our method performed better that state-of-the-art FCN-based methods.